\documentclass[sigconf]{acmart}

\usepackage{booktabs} % For formal tables
\usepackage{latexsym}
\usepackage{xcolor}

\usepackage{graphicx}  %Required
\usepackage{latexsym}
\usepackage{amsmath}
\usepackage{fancyhdr}
\usepackage{adjustbox}
\usepackage{multirow}
\usepackage{float}
\usepackage{textcomp}
\usepackage{amssymb,amsthm,bm}
\usepackage{bbm}
\DeclareMathOperator*{\argmax}{arg\,max}

\usepackage{balance}

\begin{document}

\title{Open-world Learning and Application to Product Classification}
%\titlenote{Produces the permission block, and copyright information}
%\subtitle{Extended Abstract}
%\subtitlenote{The full version of the author's guide is available as \texttt{acmart.pdf} document}

\author{Hu Xu\textsuperscript{\text{1}}, Bing Liu\textsuperscript{\text{1}}, Lei Shu\textsuperscript{\text{1}} and P. Yu\textsuperscript{\text{2}} }
\affiliation{
\textsuperscript{1}{University of Illinois at Chicago, Chicago, IL, USA}
}
\affiliation{
\textsuperscript{2}{Tsinghua University, Beijing, China}
}
\email{hxu48@uic.edu,liub@uic.edu,lshu3@uic.edu,psyu@tsinghua.edu.cn}

%\author{Hu Xu,   Bing Liu,  Lei Shu}
%\authornote{Dr.~Trovato insisted his name be first.}
%\affiliation{%
%  \institution{University of Illinois at Chicago}
  %\streetaddress{P.O. Box 1212}
%  \city{Chicago}
%  \state{IL}
%  \country{USA}
%\email{{hxu48, liub, lshu3}@uic.edu}

%\author{P. Yu}
%\affiliation{%
%    \institution{Tsinghua University}
    %\streetaddress{8600 Datapoint Drive}
%    \city{Beijing}
    %\state{Texas}
    %\postcode{78229}
%    \country{China}
%}
%\email{psyu@tsinghua.edu.cn}

%\author{Bing Liu}
%\authornote{The secretary disavows any knowledge of this author's actions.}
%\affiliation{%
%  \institution{University of Illinois at Chicago}
  %\streetaddress{P.O. Box 1212}
%  \city{Chicago}
%  \state{IL}
%  \country{USA}
  %\postcode{}
%}
%\email{liub@uic.edu}

%\author{Lei Shu}
%\authornote{This author is the one who did all the really hard work.}
%\affiliation{%
%  \institution{University of Illinois at Chicago}
  %\streetaddress{1 Th{\o}rv{\"a}ld Circle}
%  \city{Chicago}
%  \state{IL}
%  \country{USA}
% }
% \email{lshu3@uic.edu}

%\author{P. Yu}
%\affiliation{%
%  \institution{Tsinghua University}
  %\streetaddress{8600 Datapoint Drive}
%  \city{Beijing}
  %\state{Texas}
  %\postcode{78229}
%  \country{China}
%}
%\email{psyu@tsinghua.edu.cn}

\renewcommand{\shortauthors}{H. Xu et al.}

\begin{abstract}
Classic supervised learning makes the \textit{closed-world assumption} that the classes seen in testing must have appeared in training. However, this assumption is often violated in real-world applications. For example, in a social media site, new topics emerge constantly and in e-commerce, new categories of products appear daily. 
A model that cannot detect new/unseen topics or products is hard to function well in such open environments.  
A desirable model working in such environments must be able to (1) reject examples from unseen classes (not appeared in training) and (2) incrementally learn the new/unseen classes to expand the existing model. This is called \textit{open-world learning} (OWL).
This paper proposes a new OWL method based on meta-learning. 
The key novelty is that the model maintains only a dynamic set of seen classes that allows new classes to be added or deleted with no need for model re-training. 
Each class is represented by a small set of training examples. 
In testing, the meta-classifier only uses the examples of the maintained seen classes (including the newly added classes) on-the-fly for classification and rejection. 
Experimental results with e-commerce product classification show that the proposed method is highly effective\footnote{The data and code are available at \url{https://www.cs.uic.edu/~hxu/}.}.
\end{abstract}


%\ccsdesc[300]{Information systems~Clustering}


\maketitle

\section{Introduction}
An AI agent working in the real world must be able to recognize the classes of things that it has seen/learned before and detect new things that it has not seen and learn to accommodate the new things. This learning paradigm is called \textit{\underline{o}pen-\underline{w}orld \underline{l}earning} (OWL) %(or open-world recognition in computer vision) 
\cite{chen2018lifelong,bendale2015towards,fei2016learning}. 
%(? do you think we should separate open-set recognition (only do rejection forever) clearly from open world learning (allow incrementally adding or deleting) or give another new name? e.g., dynamic open world learning?)
This is in contrast with the classic supervised learning paradigm which makes the \textit{closed-world assumption} that the classes seen in testing must have appeared in training. With the ever-changing Web, the popularity of AI agents such as intelligent assistants and self-driving cars that need to face the real-world open environment with unknowns, OWL capability is crucial.  %environments and interact with humans and other systems, 
% as AI agent such as an intelligent personal assistant, chatbots, self-driving car is becoming increasingly popular. 

For example, with the growing number of products sold on Amazon from various sellers, it is necessary to have an open-world model that can automatically classify a product based on a set $S$ of product categories.
An emerging product not belonging to any existing category in $S$ should be classified as ``unseen'' rather than one from $S$.
Further, this unseen set may keep growing. When the number of products belonging to a new category is large enough, it should be added to $S$.
An open-world model should easily accommodate this addition with a low cost of training since it is impractical to retrain the model from scratch every time a new class is added.
As another example, the very first interface for many \underline{i}ntelligent \underline{p}ersonal \underline{a}ssistants (IPA) (such as Amazon Alexa, Google Assistant, and Microsoft Cortana) is to classify user utterances into existing known domain/intent classes (e.g., Alexa's skills) and also 
reject/detect utterances from unknown domain/intent classes (that are currently not supported).
% This type of classification is known as \textit{open-world classification} \cite{bendale2015towards,fei2016learning} (or open-set recognition in computer vision).
But, with the support to allow the 3rd-party to develop new skills (Apps), such IPAs must recognize new/unseen domain or intent classes and include them in the classification model\cite{kumar2017zero,kim2018efficient}. These real-life examples present a major challenge to the maintenance of the deployed model.
%The existing solution, in this case, is simply to re-train the whole model periodically \cite{kim2018efficient}. As a result, the new skills added by the 3rd-parties may not be effective until the next scheduled re-training by Amazon. 

Most existing solutions to OWL are built on top of closed-world models \cite{bendale2015towards,bendale2016towards,fei2016learning,shu-xu-liu:2017:EMNLP2017}, e.g., by setting thresholds on the logits (before the softmax/sigmoid functions) to reject unseen classes which tend to mix with existing seen classes. One major weakness of these models is that they cannot easily add new/unseen classes to the existing model without re-training or incremental training (e.g., OSDN \cite{bendale2016towards} and DOC \cite{shu-xu-liu:2017:EMNLP2017}). %and the request for adding new classes may not from a single party. (??? I added the part after and to help the following sentence)
There are incremental learning techniques (e.g., iCaRL \cite{rebuffi2017icarl} and DEN \cite{lee2017lifelong}) that can incrementally learn to classify new classes. However, they miss the capability of rejecting examples from unseen classes.
This paper proposes to solve OWL with both capabilities in a very different way via meta-learning.
%Our problem is stated as follows:   problem of (?? this name does not reflect our problem well. Need a new name.) \underline{I}ncremental \underline{O}pen-world \underline{T}ext \underline{C}lassification (IOTC), which is defined as following:\\
% (??? if we have extra space, maybe add a figure of venn diagram for this problem)

\textbf{Problem Statement}: At any point in time, the learning system is aware of a set of seen classes $S=\{c_1, \dots, c_m\}$ %($S$ can be $\varnothing$) 
and has an OWL model/classifier for $S$ but is unaware of a set of unseen classes $U=\{c_{m+1}, \dots\}$ (any class not in $S$ can be in $U$) that the model may encounter. The goal of an OWL model is two-fold: (1) classifying examples from classes in $S$ and reject examples from classes in $U$, and (2) when a new class $c_{m+1}$ (without loss of generality) is removed from $U$ (now $U=\{c_{m+2}, \dots\}$) and added to $S$ (now $S=\{c_1, \dots, c_m, c_{m+1}\}$, still being able to perform (1) without re-training the model.

Two main challenges for solving this problem are: (1) how to enable the model to classify examples of seen classes into their respective classes and also detect/reject examples of unseen classes, and (2) how to incrementally include the new/unseen classes when they have enough data without re-training the model.

As discussed above, existing methods either focus on the challenge (1) or (2), but not both. To tackle both challenges in an unified approach, this paper proposes an entirely new OWL method based on meta-learning \cite{thrun2012learning,andrychowicz2016learning,fernando2017pathnet,finn2017model,finn2018probabilistic}. The method is called \textit{\underline{L}earning to \underline{A}ccept \underline{C}lasses} (L2AC). The key novelty of L2AC is that the model maintains a dynamic set $S$ of seen classes that allow new classes to be added or deleted with no model re-training needed. Each class is represented by a small set of training examples. In testing, the meta-classifier only uses the examples of the maintained seen classes (including the newly added classes) on-the-fly for classification and rejection. That is, the learned meta-classifier classifies or rejects a test example by comparing it with its nearest examples from each seen class in $S$. Based on the comparison results, it determines whether the test example belongs to a seen class or not. If the test example is not classified as any seen class in $S$, it is rejected as unseen. Unlike existing OWL models, 
the parameters of the meta-classifier are not trained on the set of seen classes but on a large number of other classes which can share a large number of features with seen and unseen classes, and thus can work with any seen classification and unseen class rejection without re-training. 
% (? shall we add the reason somewhere? The key idea that the meta-classifier can potentially work with any class is that classes share a significant amount of features or attributes. After the meta-classifier is trained with enough classes, it is capable of detecting unseen classes.)

We can see that the proposed method works like a nearest neighbor classifier (e.g., $k$NN). However, the key difference is that we train a meta-classifier to perform both classification and rejection based on a learned metric and a learned voting mechanism. Also, 
$k$NN cannot do rejection on unseen classes. 
% The major advantage of the proposed approach is that it makes open-world learning a problem of maintaining the seen class set $S$ and the (labeled) examples in each class in $S$. Once the meta-classifier is trained, the user/system can simply add any new class with its data to the seen class set $S$ without re-training the meta-classifier. The system can still perform classification and rejection simply based on the updated $S$ using the meta-classifier. 

The main contributions of this paper are as follows. 

(1) It proposes a novel approach (called L2AC) to OWL based on meta-learning, which is very different from existing approaches. % that only tackle OWL partially. 

(2) The key advantage of L2AC is that with the meta-classifier, OWL becomes simply maintaining the seen class set $S$ because both seen class example classification and unseen class example rejection/detection are based on comparing the test example with the examples of each class in $S$. To be able to accept/classify any new class, we only need to put the class and its examples in $S$. 

The proposed approach has been evaluated on product classification and the results show its competitive performance. 

%\begin{enumerate}
%\item It proposes a novel approach to open-world learning based on meta-learning, which is entirely different from existing approaches.
%\item The key advantage of the approach is that with the meta-classifier, the open-world learning problem becomes simply maintaining the seen class set $S$ because both classification and unseen class example rejection/detection are based on comparing the test example with the examples of each class in $S$. To be able to accept/classify any new class, we only need to put the class and its examples in $S$. 
%\item The proposed approach has been experimentally evaluated and the results show its competitive performance. 
%\end{enumerate}

\section{L2AC Framework}
\begin{figure*}
\centering    
\includegraphics[width=5.5in]{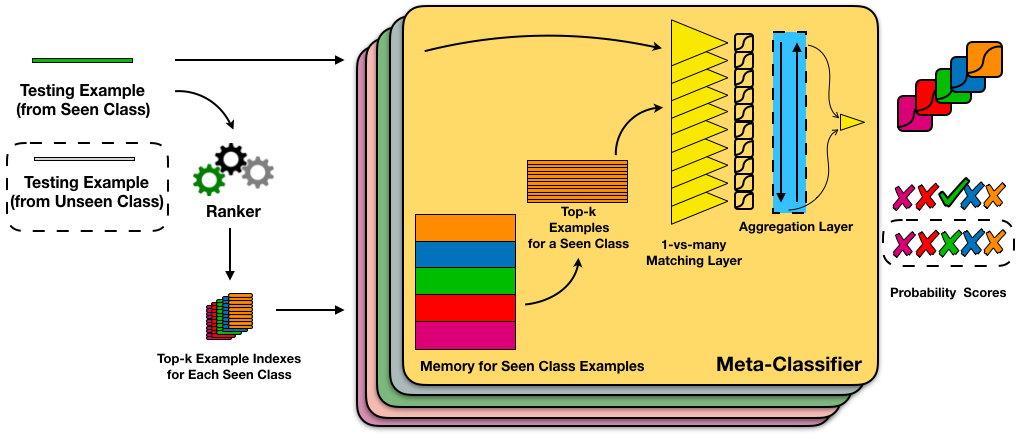}
\caption{Overview of the L2AC framework (best viewed in colors). Assume the seen class set $S$ has 5 classes and their examples are indicated by 5 different colors. L2AC has two components: a ranker and a meta-classifier. Given a (green) testing example from a seen class, the ranker first retrieves the top-$k$ nearest examples (memory indexes) from each seen class. Then the meta-classifier takes both the test example and the top-$k$ nearest examples for a seen class to produce a probability score for that class. The meta-classifier is applied 5 times (indicated by 5 rounded rectangles) over these 5 seen classes and yields 5 probability scores, where the 3rd (green) class attends the maximum score as the final class (green) prediction. However, if the test example (grey) is from an unseen class (as indicated by the dashed box), none of those probability scores from the seen classes will predict positive, which leads rejection.}
\label{fig:overview}
% \vspace{-2mm}
\end{figure*}

As an overview, Fig. \ref{fig:overview} depicts how L2AC classifies a test example into an existing seen class or rejects it as from an unseen class. The training process for the meta-classifier is not shown, which is detailed in Sec. \ref{sec:train}. 
The L2AC framework has two major components: a ranker and a meta-classifier. 
The ranker is used to retrieve some examples from a seen class that are similar/near to the test example. The meta-classifier performs classification after it reads the retrieved examples from the seen classes. The two components work together as follows.

%For a given test example (or query as it is usually called in IR), the IR model finds a list of most similar examples to from each seen class.  from seen classes to the latter for classification or rejection.
Assume we have a set of seen classes $S$. 
Given a test example %(or query as it is used to information retrieval) 
$x_t$ that may come from either a seen class or an unseen class, the ranker finds a list of top-$k$ nearest examples to $x_t$ from each seen class $c \in S$, denoted as $x_{a_{1:k}|c}$.
The meta-classifier produces the probability $p(c=1|x_t, x_{a_{1:k}|x_t,c})$ that the test $x_t$ belongs to the seen class $c$ based on $c$'s top-$k$ examples (most similar to $x_t$).
If none of these probabilities from the seen classes in $S$ exceeds a threshold (e.g., $0.5$ for the sigmoid function), L2AC decides that $x_t$ is from an unseen class (rejection); otherwise, it predicts $x_t$ as from the seen class with the highest probability (for classification). 
We denote $p(c=1|x_t, x_{a_{1:k}|x_t,c})$ as $p(c|x_t, x_{a_{1:k}})$ for brevity when necessary.
Note that although we use a threshold, this is a general threshold that is not for any specific classes as in other OWL approaches but only for the meta-classifier. More practically, this threshold is pre-determined (not empirically tuned via experiments on hyper-parameter search) and the meta-classifier is trained based on this fixed threshold. 

As we can see, the proposed framework works like a supervised lazy learning model, such as the $k$-nearest neighbor ($k$NN) classifier.
Such a lazy learning mechanism allows the dynamic maintenance of a set of seen classes, where an unseen class can be easily added to the seen class set $S$. 
However, the key differences are that all the metric space, voting and rejection are learned by the meta-classifier.
%Also, the top-$k$ most similar examples $x_{a_{1:k}|c}$ to $x_q$ is a ranked list like a IR rank list. 
%This is why we use the IR setting and refer to the first component as an ranker.
%Note the meta-classifier only focuses on the top-$k$ similar examples from each seen class for the test example $x_t$.
%(??? don't understand the small data part. the number of examples per class is fixed. So a big concept class typically has sparse examples, which may not be useful. ?? not a good argument because small data means the data is very sparse, not related to local or not local. Consider delete this part.) This allows the meta-classifier to handle classes of different sizes (or granularities) because the meta-classifier only has ``local view'' of the class. Handling classes of different sizes can be important in open-world learning because classes of different sizes are mixed together.
%For example, a big ``shopping'' domain can be mixed with a small ``weather'' domain in an intelligent personal assistant.

Retrieving the top-$k$ nearest examples $x_{a_{1:k} }$ for a given test example $x_t$ needs a ranking model (the ranker).
We will detail a sample implementation of the ranker in Sec. \ref{sec:exp} 
and discuss the details of the meta-classifier in the next section.

\subsection{Meta-Classifier}
Meta-classifier serves as the core component of the L2AC framework. It is essentially a binary classifier on a given seen class. It takes the top-$k$ nearest examples (to the test example $x_t$) of the seen class as the input and determines whether $x_t$ belongs to that seen class or not.
In this section, we first describe how to represent examples of a seen class. Then we describe how the meta-classifier processes these examples together with the test example into an overall probability score (via a voting mechanism) for deciding whether the test example should belong to any seen class (classification) or not (rejection). Along with that we also describe how a joint decision is made for open-world classification over a set of seen classes. Finally, we describe how to train the meta-classifier via another set of meta-training classes and their examples.

\subsubsection{Example Representation and Memory}
\label{sec:mem}
Representation learning lives at the heart of neural networks. 
Following the success of using pre-trained weights from large-scale image datasets (such as ImageNet \cite{russakovsky2015imagenet}) as feature encoders, we assume there is an encoder that captures almost all features for text classification.

%to the best of our knowledge, there are no similar encoders for text classification.
Given an example $x$ representing a text document (a sequence of tokens), we obtain its continuous representation (a vector) via an encoder $h=g(x)$, where the encoder $g(\cdot)$ is typically a neural network (e.g., CNN or LSTM).
We will detail a simple encoder implementation in Sec. \ref{sec:exp}.

Further, we save the continuous representations of the examples into the memory of the meta-classifier.
So later, the top-$k$ examples can be efficiently retrieved via the index (address) in the memory. 
The memory is essentially a matrix $E \in \mathbb{R} ^{n \times |h|}$, where $n$ is the number of all examples from seen classes and $|h|$ is the size of the hidden dimensions.
Note that we will still use $x$ instead of $h$ to refer to an example for brevity.
Given the test example $x_t$, the meta-classifier first looks up the actual continuous representations $x_{a_{1:k} }$ of the top-$k$ examples for a seen class. %$c$ returned by the ranker.
Then the meta-classifier computes the similarity score between $x_t$ and each $x_{a_{i} }$ ($1\le i \le k$) individually via a 1-vs-many matching layer as described next. 

\subsubsection{1-vs-many Matching Layer}
\label{sec:1vsmany}
% (? shall we mention siamese network somewhere ? I previously tend to not mention it to reduce the novelty and also -- we have almost nothing to be really "siamesed", no shared network, we compute vector in the similarity space directly, no shared network for representation learning. But one reviewer mention this paper is essentially siamese network and nothing new)
To compute the overall probability between a test example and a seen class, a 1-vs-many matching layer in the meta-classifier first computes the individual similarity score between the test example and each of the top-$k$ retrieved examples of the seen class. 
The 1-vs-many matching layer essentially consists of $k$ shared matching networks as indicated by big yellow triangles in Fig. \ref{fig:overview}.
We denote each matching network as $f(\cdot, \cdot)$ and compute similarity scores $r_{1:k}$ for all top-$k$ examples $r_{1:k}=f(x_t, x_{a_{1:k}} )$. %(?? what is $r$? matching score or similarity score).

The matching network first transforms the test example $x_t$ and $x_{a_{i}}$ from the continuous representation space to a single example in a similarity space.
% (??? I move those formulas here)
%The similarity matching layer computes the similarity score from a pair of examples (one query example $h_q$ and one candidate example $h_a$ (?? good to be explicit about where this is from and which class. The reader might have forgotten)).
We leverage two similarity functions to obtain the similarity space.
The first function is the absolute values of the element-wise subtraction:
$f_\text{abssub}(x_t, x_{a_{i}})=|x_t - x_{a_{i}}|$.
The second one is the element-wise summation:
$f_\text{sum}(x_t, x_{a_{i}})=x_t + x_{a_{i}}$.
Then the final similarity space is the concatenation of these two functions' results:
$f_\text{sim}(x_t, x_{a_{i}})=f_\text{abssub}(x_t, x_{a_{i}}) \oplus f_\text{sum}(x_t, x_{a_{i}})$, where $\oplus$ denotes the concatenation operation.
%The reason we use two similarity functions is that a single one is not sufficient to get good enough features in the similarity space.
We then pass the result to two fully-connected layers (one with Relu activation) and a sigmoid function: 
\begin{equation}
\label{eq:r}
r_i=f(x_t, x_{a_i} )=\sigma\Big(W_2\cdot\text{Relu}\big(W_1\cdot f_\text{sim}(x_t, x_{a_i} ) +b_1\big)+b_2\Big).
\end{equation}
Since there are $k$ nearest examples, we have $k$ similarity scores denoted as $r_{1:k}$.
The hyper-parameters are detailed in Sec. \ref{sec:exp}.

\subsubsection{Open-world Learning via Aggregation Layer}
\label{sec:agg}
After getting the individual similarity scores, an aggregation layer in the meta-classifier merges the $k$ similarity scores into a single probability indicating whether the test example $x_t$ belongs to the seen class.
By having the aggregation layer, the meta-classifier essentially has a \textit{parametric voting mechanism} so that it can learn how to vote on multiple nearest examples (rather than a single example) from a seen class to decide the probability.
As a result, the meta-classifier can have more reliable predictions, which is studied in Sec. \ref{sec:exp}.
%By reading these example scores in order, the meta-classifier dynamically decides the overall similarity between the test example and the top-$k$ examples from a seen class (e.g., if the top-1 is not similar enough, the top-2 may help (?? do not understand)). 
%(??? good suggestion, see if the above looks good. I think you can add something say why we need the top-k examples ranked, rather than just a simple list without ranking. Otherwise, people don't understand the purpose of ranking) 

We adopt a (many-to-one) BiLSTM \cite{hochreiter1997long,schuster1997bidirectional} as the aggregation layer.
%that can read $k$ similarity scores and make a single prediction.
We set the output size of BiLSTM to 2 (1 per direction of LSTM). 
Then the output of BiLSTM is connected to a fully-connected layer followed by a sigmoid function that outputs the probability.
The computation of the meta-classifier for a given test example $x_t$ and $x_{a_{1:k}}$ for a seen class $c$ can be summarized as: 
\begin{equation}
    \label{eq:p}
p(c|x_t, x_{a_{1:k}} )=\sigma\big(W\cdot \text{BiLSTM}(r_{1:k})+b\big).
\end{equation}
%Lastly,
Inspired by DOC \cite{shu-xu-liu:2017:EMNLP2017}, 
for each class $c \in S$, we evaluate Eq. \ref{eq:p} as:
\begin{equation} 
    \label{eq:rej}
    %\resizebox{0.42\textwidth}{!}{
        \hat{y} = \left\{
        \begin{array}{c}
        \textit{reject}, \text{ if } \max_{c \in S} p(c|x_t, x_{a_{1:k}} ) \le 0.5 ;\\
        \\
        \argmax_{c \in S} p(c|x_t, x_{a_{1:k}} ) ,\text{ otherwise.}
        \end{array} \right.
    %     }
\end{equation}
If none of existing seen classes $S$ gives a probability above $0.5$, we \emph{reject} $x_t$ as an example from some unseen class.
% (? both NIPS and AAAI reviewers ask for speed test, but without a parallel implementation (not worth to implement one), its hard to be a fair comparison. Reviewers are not easily aware that all DL libraries running with GPUs already have parallel implementation by Google or Nvidia running behind, e.g., DOC's multiple sigmoid function or parallel multi-class predictions from fully-connected layer before softmax)
Note that given a large number of classes, eq. \ref{eq:rej} can be efficiently implemented in parallel. We leave this to future work.
To make L2AC an easily accessible approach, we use $0.5$ as the threshold naturally and do not introduce an extra hyper-parameter that needs to be artificially tuned.
Note also that as discussed earlier, the seen class set $S$ and its examples can be dynamically maintained (e.g., one can add to or remove from $S$ any class). So the meta-classifier simply performs open-world classification over the current seen class set $S$.
%(??? I highlighted data and model separation is important so to have flexibility as humans, it could be a track of papers. is that too conceptual? this is an important point. We need this in the introduction as a major advantage of the new approach) 

\subsection{Training of Meta-Classifier}
\label{sec:train}
% (??? I added a reason)
Since the meta-classifier is a general classifier that is supposed to work for any class,
training the meta-classifier $p_\theta(c|x_t, x_{a_{1:k}|x_t, c} )$ % (??? one meta-classier for all seen classes: I abuse the notation here. p(c) is actually p(c=1), a binomial distribution, the condition is on candidates, not in model inputs.  //You: should the class c be included here in the condition to make it explicit. Do you produce one meta-classier for all seen classes and build one meta-classier for each seen class?) 
requires examples from another set $M$ of classes called \textit{meta-training classes}.
%We call $M$ \emph{meta-training set}.
A large $|M|$ is desirable so that meta-training classes have good coverage of features for seen and unseen classes in testing, which is in similar spirit to few-shot learning \cite{lake2011one}. 
%$|M|$ is typically very large to have a good coverage of different classes.
%(??? I remove it)
%But the number of examples for each class can be small (?? tricky to say this because we need to test it)
%This is similar 
We also enforce $ (S\cup U) \cap M=\varnothing$ in Sec. \ref{sec:exp}, so that all seen and unseen classes are totally unknown to the meta-classifier.

Next, we formulate the meta-training examples from $M$, which consist of a set of pairs 
(with positive and negative labels).
%(??? didn't understand this part: without including the class label, which is positive or negative; see below). 
The first component of a pair is a training document $x_q$ from a class in $M$, and the second component is a sequence of top-$k$ nearest examples also from a class in $M$. 

We assume every example (document) of a class in $M$ can be a training document $x_q$.
Assuming $x_q$ is from class $c \in M$,
a positive training pair is $(x_q, x_{a_{1:k}|x_q, c})$, where $x_{a_{1:k}|x_q, c}$ are top-$k$ examples from class $c$ that are most similar or nearest to $x_q$;
a negative training pair is $(x_q, x_{a_{1:k}|x_q, c'})$, where $c' \in M$, $c \neq c'$ and $x_{a_{1:k}|x_q, c'}$ are top-$k$ examples from class $c'$ that are nearest to $x_q$.
We call $c'$ one \emph{negative class} for $x_q$.
Since there are many negative classes $c' \in M\backslash c$ for $x_q$, we keep top-$n$ negative classes for each training example $x_q$. 
That is, each $x_q$ has one positive training pair and $n$ negative training pairs.
To balance the classes in the training loss, we give a weight ratio $n:1$ for a positive and a negative pair, respectively.
% We detail the 
%finding 
% (? investigation of the top-$n$ negative classes in Sec. \ref{sec:exp}.

Training the meta-classifier also requires validation classes for model selection (during optimization) and hyper-parameters ($k$ and $n$) tuning (as detailed in Experiments).
Since the classes tested by the meta-classifier are unexpected, we further use a set of  \textit{validation classes} $M'\cap M=\varnothing$ (also $M'\cap (S\cup U)=\varnothing$), to ensure generalization on the seen/unseen classes.

% (? I removed the following to save space)
%Note that the meta-training can also leverage the example indexes and memory for efficient training (to avoid loading concrete examples every time).
%But the memory must be swapped to the examples of seen classes after meta-training.

\section{Experiments}
\label{sec:exp}
%(?? needs rewrite to say we use a large product dataset, etc. and then RQ2 and RQ3) 
We want to address the following Research Questions (RQs):
%\textbf{RQ1}: (?? we cannot say this any more as we want to do product classification. We can directly say we use the Amazon data) What is the public datasets 
%that live in a dynamic world with large enough product categories 
%for product classification 
%in the domain of text classification 
%for open-world learning using meta-learning: 
\textbf{RQ1} - what is the performance of the meta-classifier with different settings of top-$k$ examples and $n$ negative classes?
\textbf{RQ2} - How is the performance of L2AC compared with state-of-the-art text classifiers for open-world classification (which all need some forms of re-training).
%open-world learning?

\subsection{Dataset}
%(?? we still need the description of the data, but we cannot talk about RQ1 any more) 
%(? I assume we do not need the following paragraph any more)
%Two datasets were used in \cite{shu-xu-liu:2017:EMNLP2017} (with the state-of-the-art text classifier for open-world learning): 20-Newsgroup (20 classes) and reviews (50 classes). They both have small numbers of classes. We also noticed that the review dataset has a large overlapping of classes, which explains the weak result of only 0.666 in the F1 score. As training a meta-classifier requires an extra meta-training set with a large number of classes, so we decide to adopt a dataset with a large number of classes. %, which is also more realistic for open-world learning.
We leverage the huge amount of product descriptions from the Amazon Datasets \cite{he2016ups} and form the OWL task as the following.
Amazon.com maintains a tree-structured category system. 
We consider each path to a leaf node as a class.
%(product type) in the category system 
%We formulate a product type classification problem based on product descriptions. 
We removed products belonging to multiple classes to ensure the classes have no overlapping.
%and keep product types with at least 100 products (examples).
This gives us 2598 classes, where 1018 classes have more than 400 products per class.
We randomly choose 1000 classes from the 1018 classes with 400 randomly selected products per class as the \textit{encoder training set};
100 classes with 150 products per class are used as the (classification) \textit{test set}, including both seen classes $S$ and unseen classes $U$;
another 1000 classes with 100 products per class are used as the \textit{meta-training set} (including both $M$ and $M'$).
For the 100 classes of the test set, we further hold out 50 examples (products) from each class as test examples. 
The rest 100 examples are training data for baselines, or seen classes examples to be read by the meta-classifier (which only reads those examples but is not trained on those examples).
To train the meta-classifier, we further split the meta-training set as 900 \textit{meta-training classes} ($M$) and 100 \textit{validation classes} ($M'$).%\footnote{We will release all selections of datasets for future research.}

%\subsection{Preprocessing}
For all datasets, we use NLTK\footnote{https://www.nltk.org/} as the tokenizer, and regard all words that appear more than once as the vocabulary.
This gives us 17,526 unique words.
%and 10,XXX words for question classification.
We take the maximum length of each document as 120 since the majority of product descriptions are under 100 words.

\subsection{Ranker}
% (? this parts need to be carefully written. both NIPS and AAAI ask for running time but without a parallel optimized implementation (not worth to implement one), it's not fair to compare.)
%Since a high-performance ranker is not our focus, 
We use cosine similarity to rank the examples in each seen (or meta-training) class for a given test (or meta-training) example $x_t$ (or $x_q$)\footnote{Given many examples to process, the ranker can be implemented in a fully parallel fashion to speed up the processing, which we leave to future work as it is beyond the scope of this work.}.
We apply cosine directly on the hidden representations of the encoder as $cosine(h_*, h_{a_{i}})=\frac{h_* \cdot h_{a_{i}}}{|h_*|_2|h_{a_{i}}|_2}$, where $*$ can be either $t$ or $q$, $|\cdot|_2$ denotes the $l$-2 norm and $\cdot$ denotes the dot product of two examples.

Training the meta-classifier also requires a ranking of negative classes for a meta-training example $x_q$, as discussed in Sec. \ref{sec:train}.
%(?? where? give a context here).
We first compute a \textit{class vector} for each meta-training class. 
This class vector is averaged over all encoded representations of examples of that class.
Then we rank classes by computing cosine similarity between the class vectors and the meta-training example $x_q$.
The top-$n$ (defined in the previous section) classes are selected as negative classes for $x_q$.
We explore different settings of $n$ later.

\subsection{Evaluation}
Similar to \cite{shu-xu-liu:2017:EMNLP2017}, we choose 25, 50, and 75 classes from the (classification) test set of 100 classes as the seen classes for three (3) experiments.
%However, these sets are incrementally sampled rather than independently sampled in \cite{shu-xu-liu:2017:EMNLP2017} (?? any advantage?). So we ensure the 25 classes are a subset of the 50 classes and 50 classes are a subset of the 75 classes.
Note that each class in the test set has 150 examples, where 100 examples are for the training of baseline methods or used as seen class examples for L2AC and 50 examples are for testing both the baselines and L2AC.
We evaluate the results on all 100 classes for those three (3) experiments.
For example, when there are 25 seen classes, testing examples from the rest 75 unseen classes are taken as from one \textit{rejection class} $c_\text{rej}$, as in \cite{shu-xu-liu:2017:EMNLP2017}.
%In this way, we mimic the real-world scenario where any unseen classes may come at any time and a small set of seen classes typically needs more rejection for unseen classes.

Besides using macro F1 as used in \cite{shu-xu-liu:2017:EMNLP2017}, we also use weighted F1 score overall classes (including seen and the rejection class) as the evaluation metric.
Weighted F1 is computed as 
\begin{equation}
\sum_{c\in S\cup\{c_\text{rej}\} } \frac{N_c}{\sum_{c\in S\cup\{c_\text{rej}\}}N_c  }\cdot \text{F1}_c,
\end{equation}
where $N_c$ is the number of examples for class $c$ and $\text{F1}_c$ is the F1 score of that class.
We use this metric because macro F1 has a bias on the importance of rejection when the seen class set is small (macro F1 treats the rejection class as equally important as one seen class).
For example, when the number of seen classes is small, the rejection class should have a higher weight as a classifier on a small seen set is more likely challenged by examples from unseen classes.
Further, to stabilize the results, we train all models with 10 different initializations and average the results.

\subsection{Hyper-parameters}
For simplicity, we leverage a BiLSTM \cite{hochreiter1997long,schuster1997bidirectional} on top of a GloVe \cite{pennington2014glove} embedding (840b.300d) layer as the encoder (other choices are also possible).
Similar to feature encoders trained from ImageNet~\cite{russakovsky2015imagenet}, we train classification over the encoder training set with 1000 classes and use 5\% of the encoding training data as encoder validation data. 
% (? I removed the hp table to save space)
%The hyper-parameters of the encoder are detailed in Table \ref{tbl:hyp}.
We apply dropout rates of 0.5 to all layers of the encoder. 
The classification accuracy of the encoder on validation data is \textbf{81.76\%}.
The matching network (the shared network within the 1-vs-many matching layer) has two fully-connected layers, where the size of the hidden dimension is 512 with a dropout rate of 0.5.
%Its hyper-parameters are also given in Table \ref{tbl:hyp}. 
We set the batch size of meta-training as 256.

%\begin{table}[t]
%    \centering
%    \scalebox{0.67}{
%        \begin{tabular}{c||c|c}
%        \hline
%        {\bf Layers} &{\bf Out Dims } &{\bf Params }  \\
%        \hline
%        Embedding & 300 & -\\
%        Dropout & 300 & 0.5 \\
%        BiLSTM & 512 & -\\
%        ReLU & 512 & -\\
%        Dropout & 512 & 0.5 \\
%        FC & 1000 & -\\
%        Softmax & 1000 & -\\
%        \hline
%        \end{tabular}
%    }
%    \quad
%    \scalebox{0.67}{
%        \begin{tabular}{c||c|c}
%        \hline
%        {\bf Layers} &{\bf Out Dims } &{\bf Params }  \\
%        \hline
%        Memory & 512 & -\\
%        AbsSub & 512 & -\\
%        Sum & 512 & - \\
%        FC & 512 & -\\
%        Dropout & 512 & 0.5 \\
%        Sigmoid & 1 & -\\
%       \hline
%       \end{tabular}
%    }
%    \caption{The hyperparameters for text encoder and matching network.}
%    \label{tbl:hyp} 
%\end{table}

To answer RQ1 on two hyper-parameters $k$ (number of nearest examples from each class) and $n$ (number of negative classes), we use the 100 validation classes to determine these two hyper-parameters.
We formulate the validation data similar to the testing experiment on 50 seen classes.
For each validation class, we select 50 examples for validation. The rest 50 examples from each validation seen class are used to find top-$k$ nearest examples.
We perform grid search of averaged weighted F1 over 10 runs for $k\in\{1, 3, 5, 10, 15, 20\}$ and $n\in \{1, 3, 5, 9\}$, where \textbf{$k=5$} and \textbf{$n=9$} reach a reasonably well weighted F1 (87.60\%). Further increasing $n$ gives limited improvements (e.g., 87.69\% for $n=14$ and 87.68\% for $n=19$, when $k=5$). But a large $n$ significantly increases the number of training examples (e.g., $n=14$ ended with more than 1 million meta-training examples) and thus training time. So we decide to select $k=5$ and $n=9$ for all ablation studies below.
Note the validation classes are also used to compute (formulated in a way similar to the meta-training classes) the validation loss for selecting the best model during Adam \cite{kingma2014adam} optimization.
%\begin{table}[t]
%    \label{tbl:hyp2} 
%    \centering

%    \caption{Hyper-parameters of matching network.}
%\end{table}

\subsection{Compared Methods}

\begin{table*}[t]
\centering    
\scalebox{0.95}{

\begin{tabular}{l||c|c|c|c|c|c}
\hline
{\bf Methods} & $|S|=25$ (WF1) & $|S|=25$ (MF1) & $|S|=50$ (WF1) & $|S|=50$ (MF1) & $|S|=75$ (WF1) & $|S|=75$ (MF1) \\
\hline
DOC-CNN & 53.25(1.0) & 55.04(0.39) & 70.57(0.46) & 76.91(0.27) & 81.16(0.47) & 86.96(0.2) \\
DOC-LSTM & 57.87(1.26) & 57.6(1.18) & 69.49(1.58) & 75.68(0.78) & 77.74(0.48) & 84.48(0.33) \\
DOC-Enc & 82.92(0.37) & 75.09(0.33) & 82.53(0.25) & 84.34(0.23) & 83.84(0.36) & 88.33(0.19) \\
\hline
DOC-CNN-Gaus & 85.72(0.43) & 76.79(0.41) & 83.33(0.31) & 83.75(0.26) & 84.21(0.12) & 87.86(0.21) \\
DOC-LSTM-Gaus & 80.31(1.73) & 70.49(1.55) & 77.49(0.74) & 79.45(0.59) & 80.65(0.51) & 85.46(0.25) \\
DOC-Enc-Gaus & 88.54(0.22) & 80.77(0.22) & 84.75(0.21) & 85.26(0.2) & 83.85(0.37) & 87.92(0.22) \\
\hline
\hline
L2AC-$n$9-NoVote & 91.1(0.17) & 82.51(0.39) & 84.91(0.16) & 83.71(0.29) & 81.41(0.54) & 85.03(0.62) \\
L2AC-$n$9-Vote3 & 91.54(0.55) & 82.42(1.29) & 84.57(0.61) & 82.7(0.95) & 80.18(1.03) & 83.52(1.14) \\
%\hline
%L2AC-Random5-$n$9 & & & \\
\hline
L2AC-$k$5-$n$9-AbsSub & 92.37(0.28) & 84.8(0.54) & 85.61(0.36) & 84.54(0.42) & 83.18(0.38) & 86.38(0.36) \\
L2AC-$k$5-$n$9-Sum & 83.95(0.52) & 70.85(0.91) & 76.09(0.36) & 75.25(0.42) & 74.12(0.51) & 78.75(0.57) \\
\hline
L2AC-$k$\textbf{5}-$n$\textbf{9} & \underline{93.07}(0.33) & 86.48(0.54) & \underline{86.5}(0.46) & 85.99(0.33) & \underline{84.68}(0.27) & 88.05(0.18) \\
L2AC-$k$5-$n$14 & \textbf{93.19}(0.19) & 86.91(0.33) & \textbf{86.63}(0.28) & 86.42(0.2) & 85.32(0.35) & 88.72(0.23) \\
L2AC-$k$5-$n$19 & 93.15(0.24) & 86.9(0.45) & 86.62(0.49) & 86.48(0.43) & \textbf{85.36}(0.66) & 88.79(0.52) \\
\hline
\end{tabular}
    }
\caption{Weighted F1 (WF1) and macro F1 (MF1) scores on a test set with 100 classes with 3 settings: 25, 50, and 75 seen classes. The set of seen classes are incrementally expanded from 25 to 75 classes (or gradually shrunk from 75 to 25 classes). The results are the averages over 10 runs with standard deviations in parenthesis.}
 \vspace{-3mm}
\label{tbl:result}
\end{table*}

\begin{figure}    
\includegraphics[width=1.55in]{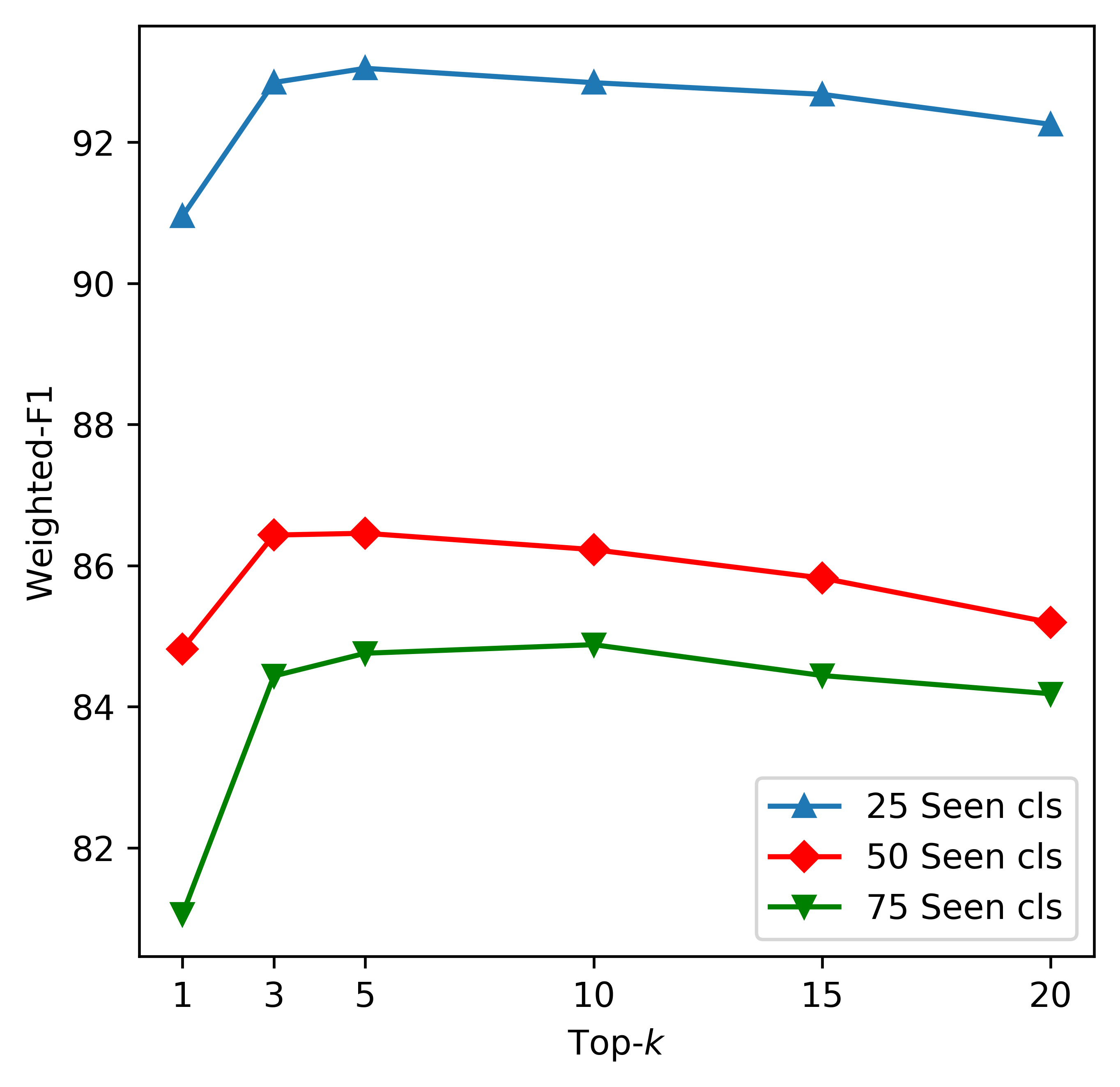}
\quad
\includegraphics[width=1.55in]{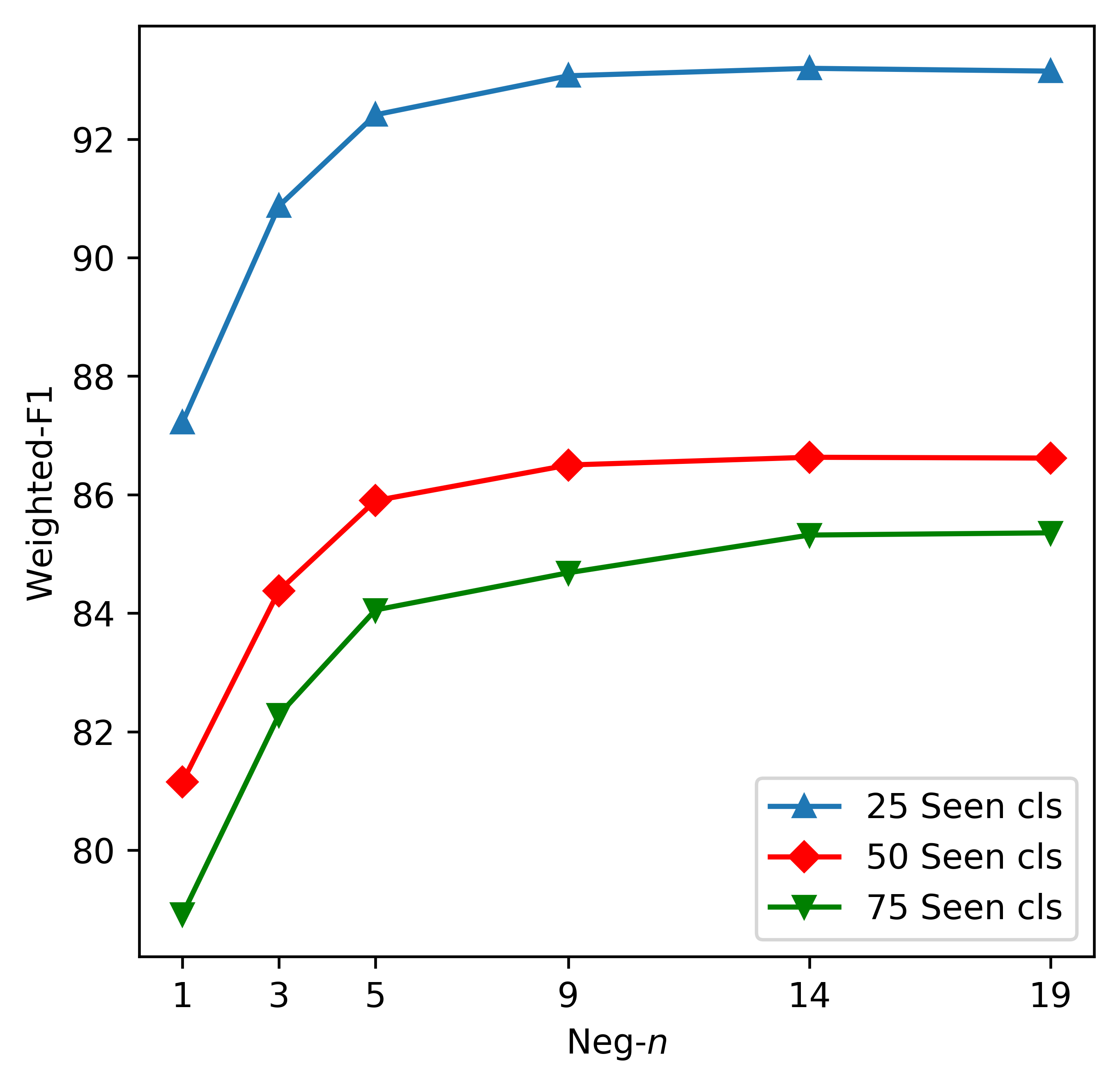}
\caption{Weighted F1 scores for different $k$'s ($n=9$) and different $n$'s ($k=5$).}
\label{fig:kn}
% \vspace{-3mm}
\end{figure}

%\textbf{Softmax}: This is a base softmax model trained on the multi-class classification task. We use this baseline as a reference baseline to show the performance of closed-world classification on the same dataset for the same architecture.
%Note that this baseline has no rejection or the capability of incremental learning.
To the best of our knowledge, DOC \cite{shu-xu-liu:2017:EMNLP2017} is the only state-of-the-art baseline for open-world learning (with rejection) for text classification. It has been shown in \cite{shu-xu-liu:2017:EMNLP2017} that DOC significantly outperforms the methods CL-cbsSVM and cbsSVM in~\cite{fei2016learning} and OpenMax in~\cite{bendale2016towards}. OpenMax is a state-of-the-art method for image classification with rejection capability. % However, DOC needs retraining from scratch to include new classes, which L2AC does not need.  
%OpenMax also was better than the method in~\cite{bendale2015towards}. 

To answer RQ2, we use DOC and its variants to show that the proposed method has comparable performance with the best open-world learning method with re-training.
Note that DOC cannot incrementally add new classes. 
So we re-train DOC over different sets of seen classes from scratch every time new classes are added to that set.
%Although 
It is thus actually unfair to compare our method with DOC because DOC is trained on the actual training examples of all classes. However, our method still performs better in general. We used the original code of DOC and created six (6) variants of it.

\vspace{+2mm}
\noindent
\textbf{DOC-CNN}: CNN implementation as in the original DOC paper without Gaussian fitting (using 0.5 as the threshold for rejection). It operates directly on a sequence of tokens. \\
\textbf{DOC-LSTM}: a variant of DOC-CNN, where we replace CNN with BiLSTM to encode the input sequence for fair comparison. BiLSTM is trainable and the input is still a sequence of tokens. \\
\textbf{DOC-Enc}: this is adapted from DOC-CNN, where we remove the feature learning part of DOC-CNN and feed the hidden representation from our encoder directly to the fully-connected layers of DOC for a fair comparison with L2AC. \\
\textbf{DOC-*-Gaus}: applying Gaussian fitting proposed in \cite{shu-xu-liu:2017:EMNLP2017} on the above three baselines, we have 3 more DOC baselines. 
Note that these 3 baselines have exactly the same models as above respectively. They only differ in the thresholds used for rejection. Gaussian fitting in \cite{shu-xu-liu:2017:EMNLP2017} is used to set a good threshold for rejection. 
We use these baselines to show that the Gaussian fitted threshold improves the rejection performance of DOC significantly but may lower the performance of seen class classification. The original DOC is \textbf{DOC-CNN-Gaus} here. 

\vspace{+1mm}
The following baselines are variants of L2AC.\\
%\textbf{L2AC-AbsSub}: we use this baseline to show that a single similarity function is not good enough. More similarity functions get better features in the similarity space.\\
\textbf{L2AC-$n$9-NoVote}: this is a variant of the proposed L2AC that only takes one most similar example (from each class), i.e., $k=1$, with one positive class paired with $n=9$ negative classes in meta-training ($n=9$ has the best performance as indicated in answering RQ1 above). 
We use this baseline to show that the performance of taking only one sample may not be good enough.
This baseline clearly does not have/need the aggregation layer and only has a single matching network in the 1-vs-many layer.\\
\textbf{L2AC-$n$9-Vote3}: this baseline uses exactly the same model as L2AC-$n$9-NoVote. But during evaluation, we allow a non-parametric voting process (like $k$NN) for prediction. We report the results of voting over top-3 examples per seen class as it has the best result (ranging from 3 to 10). If the average of the top-3 similar examples in a seen class has example scores with more than $0.5$, L2AC believes the testing example belongs to that class.
We use this baseline to show that the aggregation layer is effective in learning to vote and L2AC can use more similar examples and get better performance.\\
%\textbf{L2AC-Random5-$n$9}: this is a variant of L2AC without using the ranker (randomly select 5 examples from each class for both meta-training and testing). 
%We use this baseline to show that in open-world learning classes are in different sizes and randomly taking 5 examples from each class may introduce not-so-similar examples from a big class (?? this is arguable. We need to discuss.). \\
\textbf{L2AC-$k$5-$n$9-AbsSub/Sum}: To show that using two similarity functions ($f_\text{abssub}(\cdot, \cdot)$ and $f_\text{sum}(\cdot, \cdot)$ ) gives better results, we further perform ablation study by using only one of those similarity functions at a time, which gives us two baselines.\\
\textbf{L2AC-$k$5-$n$9/14/19}: this baseline has the best $k=5$ and $n=9$ on the validation classes, as indicated in the previous subsection. Interestingly, further increasing $k$ may reduce the performance as L2AC may focus on not-so-similar examples. We also report results on $n=14$ or $19$ to show that the results do not get much better. 

\subsection{Results Analysis}
From Table \ref{tbl:result}, we can see that L2AC outperforms DOC, especially when the number of seen classes is small. 
First, from Fig. \ref{fig:kn} we can see that $k=5$ and $n=9$ gets reasonably good results.
%(??? i forgot why i say n is small. maybe im not in status ?? what do you mean? n is already 9. how can it be small. Or do you mean n=9 is small?). 
Increasing $k$ may harm the performance as taking in more examples from a class may let L2AC focus on not-so-similar examples, which is bad for classification. More negative classes give L2AC better performance in general but further increasing $n$ beyond 9 has little impact.

% (? see the following ?? one major issue: it is clear when we incrementally adding classes how each system performs.)
%(? i guess at that time we realized L2AC is better on rejection but poor on classification. DOC is more stable but we don't wish to talk about this too much.)
Next, we can see that as we incrementally add more classes, L2AC gradually drops its performance (which is reasonable due to more classes) but it still yields better performance than DOC. Considering that L2AC needs no training with additional classes, while DOC needs full training from scratch, L2AC  represents a major advance. Note that testing on 25 seen classes is more about testing a model's rejection capability while testing on 75 seen classes is more about the classification performance of seen class examples.
From Table \ref{tbl:result}, we notice that L2AC can effectively leverage multiple nearest examples and negative classes.
In contrast, the non-parametric voting of L2AC-$n$9-Vote3 over top-3 examples may not improve the performance but introduce higher variances.
Our best $k=5$ indicates that the meta-classifier can dynamically leverage multiple nearest examples instead of solely relying on a single example.
As an ablation study on the choices of similarity functions, running L2AC on a single similarity function gives poorer results as indicated by either L2AC-$k$5-$n$9-AbsSub or L2AC-$k$5-$n$9-Sum.

DOC without encoder (DOC-CNN or DOC-LSTM) performs poorly when the number of seen classes is small.
Without Gaussian fitting, DOC's (DOC-CNN, DOC-LSTM or DOC-Enc) performance increases as more classes are added as seen classes. This is reasonable as DOC is more challenged by fewer seen training classes and more unseen classes during testing. 
As such, Gaussian fitting (DOC-*-Gaus) alleviates the weakness of DOC on a small number of seen training classes.
% (? done ?? correct? This sentence sounds weird).

%(??? good point. will remove this and always talk about OWL as a whole. now I feel the first person in this field didn't make the evaluation of open-learning problem clear enough so we somehow always ended with a mixing of closed-world and rejection. 
%I 'd more interested in train a meta-rejector (so it can be pluged with or joint tuning with any existing closed-world classifier). I feel mix those two things together as existing research could be problematic.
%I think always do a separate evaluation is very clean to show the pros and cons of each model. ?? Bing> this paragraph is tricky because no separate results for seen and unseen are given) Overall, L2AC's performance on rejection is very good. Its performance on classification is good too (??? we beat all DOC baselines but not as good as rejection ?? which table shows that?).
%But its performance decreases when more seen classes are available, which is reasonable as it is hard for the classifier with more classes. %indicating it is challenged by seen class classification. 
%But it still outperforms DOC (??? for Gaussian fitting version, NO. Gaussian fitting harm classification too ?? does DOC improve with increased seen classes? Interesting that DOC is very stable). We leave improving L2AC's classification performance to future work.
\section{Related Work}
\label{sec:rel}
%\subsection{Product Classification}

% (? the following is comment out a lot to save space)
%\subsection{Open-world Learning} % and Class Incremental Learning}
Open-world learning has been studied in text mining and computer vision (where it is called open-set recognition) \cite{bendale2015towards,chen2018lifelong,fei2016learning}. Most existing approaches focus on building a classifier that can predict examples from unseen classes into a (hidden) \textit{rejection class}.
These solutions are built on top of closed-world classification models \cite{bendale2015towards,bendale2016towards,shu-xu-liu:2017:EMNLP2017}. Since a closed-world classifier cannot detect/reject examples from unseen classes (they will be classified into some seen classes), some thresholds are used so that these closed-world models can also be used to do rejection. However, as discussed earlier, when incrementally learning new classes, they also need some form of re-training, either full re-training from scratch \cite{bendale2016towards,shu-xu-liu:2017:EMNLP2017} or partial re-training in an incremental manner \cite{bendale2015towards,fei2016learning}. 
% can easily be classified as a seen class.
%This is because these models are not originally trained for rejection, but purely trained for seen classes and they empirically reject unseen examples based on the predictions on seen classes using some thresholds. However, our meta-classifier is trained for rejection. 

Our work is also related to class incremental learning \cite{rebuffi2017icarl,rusu2016progressive,lee2017lifelong}, where new classes can be added dynamically to the classifier. For example, 
iCaRL \cite{rebuffi2017icarl} maintains some exemplary data for each class and incrementally tunes the classifier to support more new classes. However, they also require training when each new class is added. 
%our work is quite different from such incremental learning methods because incremental learning methods do not do rejection on unseen classes as we do. 
% But it still requires training and cannot reject examples from unseen classes. CL-cbsSVM \cite{fei2016learning} is another incremental learning framework for text document. It trains a 1-vs-rest SVM classifier for each newly added positive class while maintains negative classes and exemplary examples for each seen class. However, it requires training for each new class and potential re-training for old classes when the new class significantly affect any of the old classifiers.

%\subsection{Meta-learning}
%(??? reviewer argued for zero-shot learning. ?? it might be good to cite a few more papers from meta-learning as our method itself is more related to meta-learning. If space is space is needed, zero-shot learning is not so relevant and can be deleted.) 
Our work is clearly related to meta-learning (or learning to learn) \cite{thrun2012learning}, 
which turns the machine learning tasks themselves as training data to train a meta-model and has been successfully applied to many machine learning tasks lately, such as 
\cite{andrychowicz2016learning,fernando2017pathnet,finn2017model,finn2018probabilistic,fan2018learning}.
%. For example, it has been used to learn an optimizer \cite{andrychowicz2016learning}, to learn network configurations \cite{fernando2017pathnet}, to learn initial and easy-to-tune weights for few-shot learning \cite{finn2017model,finn2018probabilistic} and to learn a teacher model that can guide training sample selection \cite{fan2018learning}.
Our proposed framework focuses on learning the similarity between an example and an arbitrary class %via reading that class' examples. %??? I feel this statement is somehow dangerous which has not been done by other meta-learning methods. 
and we are not aware of any open-world learning work based on meta-learning. 

%\subsection{Zero-shot Learning}
The proposed framework is also related to zero-shot learning \cite{lampert2009learning,palatucci2009zero,socher2013zero} (in that we do not require training but need to read training examples), 
%However, existing zero-shot learning methods mostly focus on learning a mapping from the input space to an attribute or embedding space and then using some external knowledge to make the class prediction from the attribute or embedding space. We focus on learning a generalized mapping of the input space to a binary space, so unexpected classes can also benefit from such a mapping without the requirement for training data.
%\subsection{$k$NN and Metric Learning}
%The proposed model is related to 
$k$-nearest neighbors ($k$NN) (with additional rejection capability, metric learning \cite{xing2003distance} and learning to vote), 
%as well, which also does not require any training but only requires training examples for each seen class during testing. 
%However, $k$NN is a non-parametric model that leverages a pre-defined metric to compare similarities of testing examples with training examples and uses a non-parametric voting process for classification. 
%Also, $k$NN is only used for closed-world classification and does not perform rejection.
%In contrast, our framework has a parametric model and it learns a similarity metric, a voting mechanism, and a rejection capability.
%Learning a similarity metric is related 
%and metric learning \cite{}.
and Siamese networks \cite{bromley1994signature,koch2015siamese,vinyals2016matching} (regarding processing a pair of examples). 
%Siamese networks is commonly used for few-shot classification \cite{koch2015siamese,vinyals2016matching}. 
However, all those techniques work in closed-worlds with no rejection capability.
%However, the proposed framework focuses on the metric learning part instead of the representation learning part.
%We encode all training examples into a memory instead of using raw text representation and do forward passing.
%The proposed meta-classifier only focus on similarity metric learning and voting for top-k nearest examples for each class. 
% (? may remove this cv app as we are not targetting NIPS any more depends on space.)
% From the application side, our work is related to face recognition \cite{taigman2014deepface,schroff2015facenet} in computer vision.
%However, face recognition is a controlled application, where the type (face) of all classes is pre-defined (so the variance among classes is limited and different classes share a significant amount of features (e.g., glasses, shirts, etc.) ). 
%The training data for face recognition is close to few-shot learning where the number of classes is huge but each class has only 2 or a few examples.
% However, open-world learning is generally more challenging given no restrictions on the type of class.

%\subsection{Memory Augmented Neural Networks}
%Since the proposed meta-classifier reads examples from a seen class, it is thus related to memory augmented neural networks, such as Neural Turing Machine \cite{graves2014neural} and Memory Networks \cite{sukhbaatar2015end}. 
%But we focus on building a meta-classifier that reads seen class examples to accept or reject a class.

Product classification has been studied in \cite{shen2011item,shen2012large,chen2013cost,gupta2016product,cevahir2016large,kozareva2015everyone}, mostly in a multi-level (or hierarchical) setting. % \cite{kozareva2015everyone} showed that neural network embedding representation performs well on over 300 categories in their category taxonomy.
However, given the dynamic taxonomy in nature, product classification has not been studied as an open-world learning problem.

\section{Conclusions}
In this paper, we proposed a meta-learning framework called L2AC for open-world learning. L2AC has been applied to product classification. 
Compared to traditional closed-world classifiers, our meta-classifier can incrementally accept new classes by simply adding new class examples without re-training.
Compared to other open-world learning methods, the rejection capability of L2AC is trained rather than realized using some empirically set thresholds. Our experiments showed superior performances to strong baselines.

\section*{Acknowledgments} Bing Liu's work was partially supported by the National Science Foundation (NSF IIS 1838770) and by a research gift from Huawei.

\bibliographystyle{ACM-Reference-Format}
\balance 
\bibliography{main}

\end{document}